\documentclass[10pt,a4paper]{article}
\usepackage{epsfig}
\title{\LARGE\bf Norm Based Causal Reasoning in Textual Corpus}
\author {{\sc Farid Nouioua}   {\sc } \\
\small Laboratoire d'informatique de Paris-Nord\\ 
\small {Avenue Jean-Baptiste Cl\'ement, F-93430 Villetaneuse. France}\\
\small {E-mail : nouiouaf@lipn.univ-paris13.fr}\\}

\date{}

\begin{document}

\maketitle

\section{Motivation}

Truth based entailments are not sufficient for a good comprehension of NL. In fact, it can not deduce implicit information necessary to understand a text. On the other hand, norm based entailments are able to reach this goal. Let us consider this text \cite{Kayser:04a}\cite{Kayser:04b}: "the vehicle in front of me braked". Using a truth based approach; we can obtain all the logical consequences of a formula such as: $(\exists v, t)~ Vehicle (v) \wedge Instant (t) \wedge In-Front-Of (v,~ 'me',~ t) \wedge break (v, ~t).$ While norms provide further conclusions like: $v$ and me were in the same direction, no vehicle was between $v$ and me, I had to brake when $v$ braked \ldots\\
This idea was behind the development of Frames \cite{Minsky:75} and Scripts \cite{Schank:77}\cite{Schank:79} in the 70's. But these theories are not formalized enough and their adaptation to new situations is far from being obvious.\\
Actually, no repository of norms is available for a given domain. Moreover, norms are seldom made explicit in texts, because as Schank noticed, texts do not describe the normal course of events but focus rather on the description of abnormal situations. The motivation of the present work is to extract norms by detecting their violations in the texts.\\
We are working on a corpus of 60 texts describing car crashes. For each text, we are searching the cause of the accident as perceived by a standard reader. We hypothesize that the perceived cause of an abnormal event is the violation of a norm (anomaly). Among all the anomalies evoked by a text, one of them is considered as 'primary'. It represents the most plausible cause of the accident. The other anomalies result from the primary one and are called derived anomalies.
\section{The representation language} 
In this work, we use a first order logic (FOL) representation language which takes into account modalities, time and non monotonicity.We only give here its main principles (see  \cite{Kayser:04b} for details):\\
In order to quantify over properties, we use the technique of reification commonly used in AI: a binary predicate $P(X, ~Y)$ is written: $Holds(P, ~X,~ Y)$\footnote{A predicate $Q(X,~ Y,~ Z)$ is written : $Holds(Combine(Q, ~X),~ Y,~ t)$. $Combine(X,~ t)$ is a complex property}.\\
We decompose the scene into a succession of discrete states. Each one is characterized by a set of literals keeping a stable truth value. Thus, what we represent explicitly is the linear time of the events, as they really occurred. The form of the predicates is: $Holds (P, A, t)$ where $P$ is a property, $A$ is an agent and $t$ is a state number.\\
In addition to truth-values, the texts introduce modalities. The technique of reification enables us to treat modalities as first order predicates. In our work, we use two main modalities: The first one is a kind of necessity. It expresses agent duties. $Must (P, ~A,~ t)$ means that at state $t$, agent $A$ must reach the property $P$. The second one is a kind of possibility. It expresses the capacities of the agents. $Able-To (P, ~A,~ T)$ means that at state $t$, agent $A$ is able to reach the property $P$.\\
Inference rules are written in Reiter's default logic \cite{Reiter:80}. Material implications are written $(A \rightarrow B)$, normal defaults $\frac{A:B}{B}$ are written for short $A : B$ and semi-normal ones $\frac{A:B\wedge C}{C}$ are written A : B[C].\\
A primary anomaly comes under one of the two following forms:\begin{center}$Must(P, ~A,~ t) \wedge Able-To(P,~ A,~ t) \wedge Holds(P',~ A,~ t+1) \wedge Incompatible(P,~ P')\rightarrow An$
$Holds(Combine(Disruptive\_Factor,~ C),~ A,~ t) \rightarrow An$\end{center}
Derived anomalies are expressed by:
\begin{center}$Must(P,~ A,~ t) \wedge \neg Able-To(P,~ A,~ t) \wedge Holds(P',~ A,~ t+1) \wedge Incompatible(P,~ P')\rightarrow D-An$\end{center}

\section{From the text to the cause of the accident} 

\begin{figure}[htbp]
\centering{\psfig{figure=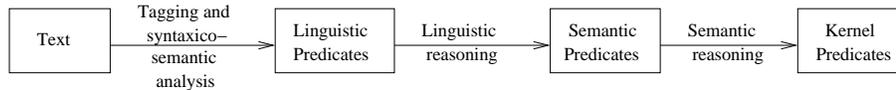,width=12cm}}
\caption{The overall architecture of the reasoning system}
\label{archi}
\end{figure}

\subsection{From the text to the semantic predicates}
First of all, a tagger and a syntactico-semantic analysis are applied to the text. The result of this step is a set of linguistic relations between relevant words of the text. The list of this type of relations, called linguistic predicates, is very short, namely:\\
$Subject(V,~ N), Object(V,~ N)$ : $N$ is the subject (resp. object) of the verb $V$.\\
$Qualif-N(N,~ A), Qualif-V(V,~ A)$ : $A$ is a qualification for the noun $N$ (resp. the verb $V$). It is the case for example for adjectives and adverbs.\\
$Compl-N(X, ~N,~ Z), Compl-V(X,~ V,~ Z)$ : $Z$ is a complement for the noun $N$ (resp. the verb $V$). It is introduced by $X$.\\
$Support(X,~ Y)$ : $X$ is a support for $Y$. For instance in "A est venu heurter B"(A comes to run up against B), we have the relation $Support(venir,~ heurter)$. Let us consider the text : \\
$"J'\acute{e}tais~\grave{a}~ l'arr\hat{e}t~ au~ feu~ rouge~ lorsque~ le~ v\acute{e}hicule~ A~ m'a~ percut\acute{e}~\grave{a}~ l'arri\grave{e}re"~$(I was stopped on a red light, when vehicle A bumped on me with the back). We obtain :
\begin{center}$Subject(\hat{e}tre, J'),~ Compl-V(\grave{a}, \hat{e}tre, arr\hat{e}t),~ Compl-N(\grave{a}, arr\hat{e}t, feu),~ Qualif-N(feu, rouge),$
$Compl-V(lorsque,~ \hat{e}tre,~percuter),~ Subject(percuter,~v\acute{e}hicule),~Qualif-N(v\acute{e}hicule,~ A),$
$Object(percuter,~ m'),~ Compl-V(\grave{a},~ percuter, ~arri\grave{e}re).$\end{center} 
After that, a non monotonic linguistic reasoning process transforms the linguistic predicates into a set of semantic predicates. For our example we obtain here:
\begin{center} $Holds(Stop,~B,~1), ~Holds(Combine(Light,~Red),~A,~1),$
$Holds(Combine(Bump,~B),~A,~2),~Holds(Combine(Shock,~ back),~B,~2)$.\end{center}
\subsection{From the semantic predicates to the kernel}
Semantic predicates are supposed to represent the explicit semantic content of the text. The semantic reasoning process uses inference rules to enrich the set of semantic predicates extracted from a text by adding further implicit conclusions. The inference rules are based on our common knowledge about the norms of the domain of car crashes.\\
The kernel contains six (reified) predicates: 
\begin{center} $Stop,~Control,~Run\_Slowly\_Enough,~Start,~Move\_Back ,~ Combine(Disruptive\_Factor,C)$.\end{center}
Computing extensions of a first order semi-normal default theory is intractable in the general case. To overcome this difficulty, one has to consider sub-sets of the theory in which some constraints must be satisfied. In the present work, predicates and rules are designed so that they can be stratified i.e. organized in layers such that the derivation of a predicate belonging to a given layer depends only on the upper layers. Formally, the stratification constraints is verified if : (L(P) denotes the number of the layer containing the predicate P).\\
Each implication $A \rightarrow B$, (resp. a normal default $A : B$) verifies : $L(A) ~\geq ~L(B)$.\\
Semi normal defaults $A : B [C]$ verify : $L(A)~ \geq~ L(B)$ and $L(C)~ > ~Max (L(A),~ L(B))$\\
Notice that rules belonging to the layer of number L(B) are those having B as conclusion.\\
In our system, the stratification is applied in two levels. In the first level the stratification is based on the modalities.Four layers are identified. The first layer contains predicates with the empty modality ($Holds(P,~ A,~ t)$). The second one is constituted by duty predicates ($Must(P,~ A,~ t)$). In the third one we find predicates of capacity ($Able-To(P,~ A,~t)$). Finally the last layer contains the two predicates $An$ (primary anomaly) and $D-An$ (derived anomaly).\\
The second level of stratification concerns the predicates of the two first layers (corresponding to empty and duty modalities). In each of these layers, we establish an order to the predicates so that the constraints of stratification are verified. We obtained 10 sub-layers in the first layer, and 2 sub-layers in the second one. We have checked manually the validity of this method on a significant part of the corpus, and we are developing an automatic reasoning system to validate automatically the obtained results.\\
Considering our example, in the first layer, the involved predicates are ordered in the following 8 sub layers (two sub layers are not used in this example): 
\begin{center}$\{Bump, ~Shock\},~\{Shock\},~ \{Stop, ~Avoid, ~Obstacle\},\{Control\},~\{predictable\},$
$\{Same\_File, ~Follow\},~ \{Stop\_Cause\},~ \{Cause\_Later\_Stop\}$.\end{center}
Among others, the following inference rules of this layer are applied:
\begin{center}$Holds(Combine(Bump, ~V), ~W,~ t) ~\rightarrow \neg Holds(Stop,~ W,~ t)~~ (sub layer~ 3).$\end{center}
\begin{center}$Holds(Combine(Shock,~V),~ W,~ t) ~\wedge~ Holds(Combine( Shock\_Pos, ~V), ~Back,~ t) : $
$Holds(Combine(Follow, ~V),~ W, ~t-1)[Holds(Control,~W,~t-1)]~~ (sub layer~ 6).$\end{center}
The first rule means that if $V$ bumps into $W$ at state $t$, then $V$ is not stopped at this state $t$. It enables us to infer $\neg Holds(Stop, ~A, ~2) (V = B,~ W = A,~ t = 2)$. The second rule states that, in general, if there is a shock between $V$ and $W$ at state $t$, and the position of the shock of $V$ is its back, then generally $W$ was following $V$ in the same file. From this rule we can obtain $Holds(Combine(Follow,~ B),~ A,~ 2) (V = B,~ W = A,~ t = 2)$. \\
We deduce: $Holds(Combine(Follow,~B),~ A,~ 1)$ by applying the backward persistence rule for the predicate Follow:
\begin{center}$Holds(Combine(Follow,~ V),~ W,~ t) : Holds(Combine(Follow,~ V),~ W,~ t-1)$\end{center}
An example of inferring duties in layer 2, is the rule :
\begin{center}$Holds(Combine(Follow,~ V),~ W,~ t)~ \wedge ~Holds(Stop,~ V,~ t)~ \rightarrow ~Must(Stop,~ W,~ t)~~ (sub layer 2)$\end{center}If $W$ follows $V$ in a file at state $t$, and $V$ stops at this state, then $W$ must stops at state $t$ too. From this rule, we infer: $Must(Stop,~ A,~ 1) (V = B,~ W = A,~ t = 1)$.
Finally to determine if $A$ is able or not to avoid the shock, use the general rule \cite{Kayser:04b}:
\begin{center}$Able-To(P,~ V,~ t)~ \leftrightarrow~ (\exists Act)~ Action(Act)~ \wedge~ Pcb(P,~ Act)~ wedge~ Available(Act,~ P,~ V,~ t)$\end{center}
At state $t$, $V$ is able to reach $P$ if and only if there is an action $Act$ which is a potential cause for $P (Pcb(P,~ Act))$ and $Act$ is available for $V$ to reach $P$ at state $t$. Knowing the fact $Pcb(Brake,~ Stop)$ and that by default any potential cause of an effect is available, we obtain $Available(Brake,~ Stop,~ A,~ 1)$. Consequently we deduce $Able-To(Stop,~ A,~ 1)$.\\
Finally, by applying the first form of primary anomalies, we can detect the cause of the accident: 
\begin{center}" A did not stop in a situation in which it had to do. "\end{center}
\section {Conclusion and Perspectives}
In the present work, we propose a norm based reasoning system able to detect the causes of the accidents from their textual descriptions. The cause is seen as a violation of a norm. The study we have done enabled us to determine a limited number of semantic predicates (50) and inference rules (currently 150). In a short and medium term perspective, we will finish the implementation of the automatic reasoning system based on the idea of stratification, complete the design of remaining inference rules and validate our approach on new car crash reports. In a longer term perspective, we will try to generalize our methodology to other domains and we will explore the possibility of applying our approach to propose a norm base indexation of textual documents.

\bibliography{/export/home/LIPN-SF1/users/RCLN/nouiouaf/articlelatex/biblio} 

\begin{thebibliography}{1}

\bibitem{Kayser:04a}
{ Kayser, D and Nouioua, F }.
\newblock { About Norms and Causes }.
\newblock In {\em { Proceedings/Actes, 17th FLAIRS Conference }}, volume~2,
  pages 502--507, 2004.

\bibitem{Kayser:04b}
{ Kayser, D and Nouioua, F }.
\newblock { Representing knowledge About Norms }.
\newblock In {\em { Proceedings/Actes, 16th ECAI Conference }}, pages 363--367,
  2004.

\bibitem{Minsky:75}
{ Minsky, M }.
\newblock { A Framework for Representing Knowledge}.
\newblock {\em { Psychology of Computer Vision (P.H. Winston, ed.) }}, pages
  211--277, 1975.

\bibitem{Reiter:80}
{ Reiter, T }.
\newblock { A Logic for Default Reasoning }.
\newblock {\em { Artificial Intelligence. Special Issue on Non Monotonic
  Logics}}, 13(1-2):81--132, 1980.

\bibitem{Schank:79}
{ Schank, R.C }.
\newblock { Interestingness: Controlling Inferences}.
\newblock {\em { Artificial Intelligence }}, 12(3):273--297, 1979.

\bibitem{Schank:77}
{ Schank, R.C and Abelson, R.P }.
\newblock { Scripts, Plans, Goals and Understanding. Lawrence Erlbaum Ass}.
\newblock 1977.

\end{thebibliography}

\end{document}